\documentclass[letterpaper,10pt,conference]{templates/ieeeconf}

\IEEEoverridecommandlockouts

\usepackage{amsmath} 
\usepackage{amssymb} 

\usepackage{graphicx} 
\usepackage{balance} 
\usepackage{amsfonts} 
\usepackage{cite} 
\usepackage[table]{xcolor} 
\usepackage{bm} 
\usepackage[hyphens]{url} 
\usepackage[hidelinks]{hyperref} 
\usepackage[switch]{lineno} 
\usepackage{tabularx} 

\newcolumntype{C}{>{\centering\arraybackslash}X}
\newcolumntype{L}{>{\raggedright\arraybackslash}X}
\newcolumntype{R}{>{\raggedleft\arraybackslash}X}

\DeclareMathOperator*{\argmax}{argmax}

\makeatletter
\let\MYcaption\@makecaption
\makeatother

\usepackage{subcaption}
\makeatletter
\let\@makecaption\MYcaption
\makeatother

\title{\LARGE\bf
    Object Instance Retrieval in Assistive Robotics: Leveraging Fine-Tuned\\
    SimSiam with Multi-View Images Based on 3D Semantic Map
}

\author{
    Taichi Sakaguchi$^{1}$
    Akira Taniguchi$^{1, *}$,
    Yoshinobu Hagiwara$^{1,2}$,
    Lotfi El Hafi$^{1}$,\\
    Shoichi Hasegawa$^{1}$, and
    Tadahiro Taniguchi$^{1,3}$
    \thanks{
        This work was supported by JSPS KAKENHI Grants-in-Aid for Scientific Research (Grant Numbers JP23K16975, 22K12212) and JST Moonshot Research \& Development Program (Grant Number JPMJMS2011).
    }
    \thanks{
        $^{1}$Taichi Sakaguchi, Akira Taniguchi, Yoshinobu Hagiwara, Lotfi El Hafi, Shoichi Hasegawa, and Tadahiro Taniguchi are with Ritsumeikan University;
        2-150 Iwakura, Ibaraki, Osaka 567-8570, Japan.
        $^{2}$Yoshinobu Hagiwara is with Soka University;
        1-236 Tangi, Hachioji, Tokyo 192-8577, Japan.
        $^{3}$Tadahiro Taniguchi is with Kyoto University;
        36-1 Yoshida-Honmachi, Sakyo, Kyoto 606-8317, Japan.
    }
    \thanks{
        $^{*}$Corresponding author. (e-mail: {\tt akira-t@fc.ritsumei.ac.jp})
    }
}

\begin{document}


\maketitle

\thispagestyle{empty}
\pagestyle{empty}



\begin{abstract}
    Robots that assist humans in their daily lives should be able to locate specific instances of objects in an environment that match a user's desired objects.
    This task is known as instance-specific image goal navigation (InstanceImageNav), which requires a model that can distinguish different instances of an object within the same class. 
    A significant challenge in robotics is that when a robot observes the same object from various 3D viewpoints, its appearance may differ significantly, making it difficult to recognize and locate accurately.
    In this paper, we introduce a method called SimView, which leverages multi-view images based on a 3D semantic map of an environment and self-supervised learning using SimSiam to train an instance-identification model on-site. 
    The effectiveness of our approach was validated using a photorealistic simulator, Habitat Matterport 3D, created by scanning actual home environments.
    Our results demonstrate a 1.7-fold improvement in task accuracy compared with contrastive language-image pre-training (CLIP), a pre-trained multimodal contrastive learning method for object searching.
    This improvement highlights the benefits of our proposed fine-tuning method in enhancing the performance of assistive robots in InstanceImageNav tasks.
    The project website is \href{https://emergentsystemlabstudent.github.io/MultiViewRetrieve/}{\textcolor{blue}{https://emergentsystemlabstudent.github.io/MultiViewRetrieve/}}.
\end{abstract}





\vspace{-4pt}
\section{Introduction}
\label{sec:introduction}

\begin{figure}[t]
    \centering
    \includegraphics[width=0.94\linewidth]{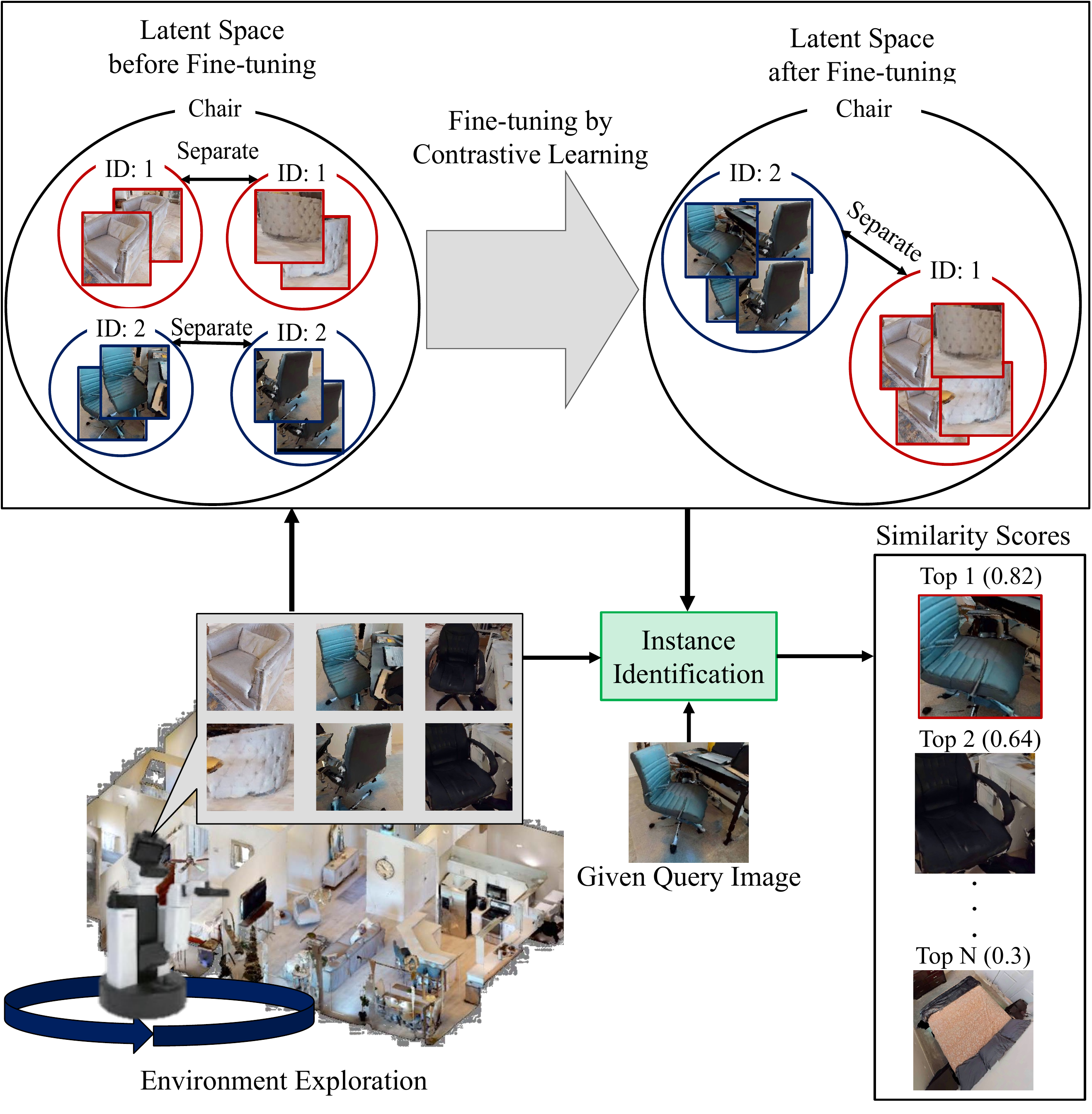}
    \caption{Tasks and challenges addressed by this study. 
    The robot observes the same object from various angles of view.
    Therefore, a model that learns the similarities between images from different 3D viewpoints is required to identify objects similar to a given query image.}
    \label{fig:concept-image}
\end{figure}

In indoor environments, when a user does not know the location of a desired object, a robot's ability to discover an object identical to a query image previously captured by the user becomes an important task.
For example, if a user wants the robot to check whether their cell phone is on a chair, the robot must move closer to the chair. 
However, multiple ``chairs'' exist in the environment, as shown in Fig.~\ref{fig:concept-image}. 
To efficiently execute tasks in such situations, the robot must distinguish between instances of the same class of objects present in the environment, identifying those desired by the user and those that are not relevant.
\textbf{Instance-specific image goal navigation (InstanceImageNav)} was proposed as a task for locating a specific instance to a query image within an environment~\cite{instancenav}. 
When a robot observes an object while exploring a 3D space, images of the same instance can include images from various 3D viewpoints, such as observing it from the back, as shown in Fig.~\ref{fig:concept-image} (top).
Here, the robot may have difficulty recognizing the two as the same instance.
Thus, a challenge is calculating the similarity of images of the same object from different 3D viewpoints.


Contrastive language--image pre-training (CLIP)~\cite{clip} has attracted interest for its applications in various robotic tasks, such as NLMap~\cite{nlmap} and CLIP-Fields~\cite{clip_fields}.
CLIP pre-trains via multimodal contrastive learning with pairs of images and their captions.
%
CLIP models effectively group objects with the same name into categories, resulting in excellent performance in category-based tasks. 
However, this approach can be ineffective in instance-based fine-grained tasks such as InstanceImageNav (see Section~\ref{sec:problem:clip}). 
This is because categories ignore the differences between instances within the same category.
Therefore, we had key insights into using unimodal contrastive learning with image pairs to address this challenge and improve the instance-based performance.

This study explored methods that pre-train image pairs through contrastive learning~\cite{simclr, simsiam, dinov2, moco} (see Section~\ref{sec:related:contrastive} for details). 
Self-distillation methods such as SimSiam~\cite{simsiam} and DINOv2~\cite{dinov2} are used to create feature vectors of images generated through data augmentation from a single image. 
Models trained through unimodal contrastive learning between images can achieve higher similarity in feature vectors for images of the same instance and acquire discriminative representations between instances~\cite{dinov2}.
Therefore, for the InstanceImageNav task, we consider the feature vector obtained from an image encoder trained through self-distillation to be more effective than those obtained from the CLIP image encoder. 
The proposed system records the feature vectors of an instance trained through contrastive learning and their positions on a map. 
We verified that self-distillation has a higher success rate than the CLIP image encoder in the InstanceImageNav task.

%
An important approach in this study was the fine-tuning of the latent space such that the feature vectors of multi-view images of the same instance become similar (see Fig.~\ref{fig:concept-image} (top) and Section~\ref{sec:problem:multi-view}). 
Robots can explore independently, collect datasets of their environment, and learn.
The proposed system automatically labels images using a robot's 3D semantic map.
This approach improves the similarity between multi-view images of the same object, enabling the simultaneous performance of contrastive learning and image-label classification within a self-supervised learning framework.
Such an approach can be applied not only to home environments but also to image searches for products in convenience stores and warehouses.

The aims of this study were as follows:
(i) To investigate whether an image encoder trained with contrastive learning~\cite{simclr, simsiam, dinov2} is more suitable for instanceImageNav than CLIP models trained with multimodal contrastive learning.
(ii) To investigate whether self-supervised learning with observed object images and pseudo-labels based on 3D semantic maps can acquire instance-specific multiview representations.

The following are the main contributions of this study:
\begin{enumerate}
    \item We show that models from contrastive learning between images are superior to CLIP, which involves multimodal contrastive learning between vision and language, for instance-level object identification.
    \item We show that increasing the similarity between identical object images from different viewpoints observed by the robot in the framework of self-supervised learning improves object instance retrieval that is identical to the query image.
\end{enumerate}

\section{Problem Statement}
\label{sec:problem}

This study addressed two critical challenges essential for advancing the InstanceImageNav using contrastive learning.

\subsection{Problem 1: Zero-shot InstanceImageNav by Contrastive Learning}
\label{sec:problem:clip}

%
CLIP solves zero-shot classification tasks with high accuracy by using similarities between object class names and image feature vectors~\cite{clip}.
However, CLIP has been suggested to be unsuitable to fine-grained tasks~\cite{birdsnap, aircraft, flowers} that search for something in sub-classes within the same class. 
%
Hence, we hypothesize that an encoder trained through self-distillation is more effective than the CLIP image encoder for the InstanceImageNav task.


\subsection{Problem 2: Visual Feature Similarity in Multi-View Images}
\label{sec:problem:multi-view}

When a robot explores an environment and observes an object, it does so from various 3D viewpoints.
Here, a problem occurs in that the success rate of InstanceImageNav decreases owing to the possibility that the appearance may be significantly different.
In addition, maximizing the similarity between images using contrastive learning with such a dataset of multi-view images may be difficult, as shown in Fig.~\ref{fig:concept-image} (top left)).
Generally, contrastive learning uses data augmentation to maximize the similarity between two different images in which a two-dimensional viewpoint change is performed on a single image.
%
To improve instance identification from different viewpoints, a pre-trained model must be fine-tuned on multi-view images of the same instance collected by the robot, thereby enhancing the similarity between the images.

%


\section{Related Studies}
\label{sec:related}

This section summarizes the related studies on the two aspects of contrastive learning and object retrieval.

\subsection{Contrastive Learning}
\label{sec:related:contrastive}

Contrastive learning is a method of learning image feature representations in a self-supervised manner.
Contrastive learning methods are broadly divided into methods that use negative pairs~\cite{simclr, moco} and self-distillation methods, which do not use negative pairs~\cite{simsiam, dino, dinov2, cookbook_ssl}.
Methods using negative pairs are trained using data augmentation to maximize the similarity between the same images and minimize the similarity between different images.
Because training is conducted to maximize the similarity between images of the same instances, a model that has learned image feature representations through contrastive learning can learn discriminative representations between different instances~\cite{reverse_engineering_ssl}. 
Therefore, models pre-trained using contrastive learning are suitable for tasks requiring discrimination between instances, such as InstanceImageNav~\cite{instancenav}.

As a self-distillation method, SimSiam was proposed as a method in which two different images generated from a single image through data augmentation are input into a convolutional neural network to maximize the cosine similarity of their feature vectors~\cite{simsiam}. 
Furthermore, DINOv2 inputs images generated through data augmentation into separate vision transformers (ViTs) to minimize the cross-entropy between the class and patch tokens of two ViTs~\cite{dinov2}.
%
Additionally, Oquab~\textit{et~al.} experimentally demonstrated a self-distillation method that achieved high accuracy in similar-image retrieval from a database of images of objects identical to those in a query image~\cite{dinov2}. 


\subsection{Object Retrieval}
\label{sec:related:search}


To solve InstanceImageNav, Kranz~\textit{et~al.} proposed a method in which navigation policies are learned through deep reinforcement learning and the robot does not have an explicit environment map~\cite{instancenav}.
However, this approach requires large amounts of data and suffers from over-fitting. 


In addition to the methods for solving object search tasks based on deep reinforcement or imitation learning~\cite{instancenav, ovrl}, methods that utilize the explicit environment map of a robot have been proposed~\cite{nlmap, clip_fields}.
For example, Chen~\textit{et~al.} proposed a method for calculating the feature vector of an observed object image using a CLIP image encoder. They constructed a semantic map called NLMap that records the position information of an object feature vector~\cite{nlmap}.
Specifically, instructions such as ``Bring me a snack'' are given.
The word feature vector of the target object ``snack'' is calculated using a CLIP text encoder.
The object search task is solved by calculating the similarity between the object feature vectors recorded in NLMap and the word feature vector and identifying the position of the most similar object.
%
A method in which a robot uses an explicit map of the environment for object search tasks was shown to achieve a higher success rate in real-world tasks than a method in which the robot learns a navigation policy in an end-to-end manner. \cite{realworld_objectnav}.
Therefore, this study adopted an approach using an explicit environment map inspired by NLMap~\cite{nlmap}.

In addition, CLIP-Fields constructs a 3D semantic map by learning, such that the feature vectors of each point on the 3D point cloud map are similar to those obtained from the CLIP image encoder~\cite{clip_fields}.
CLIP-Fields solves the object search task by calculating the similarity between the feature vector obtained from the CLIP text encoder and the feature vector of the point cloud of the 3D semantic map and then searching for the target object's position.

Semantic map methods using CLIP feature vectors, such as NLMap~\cite{nlmap} and CLIP-Fields~\cite{clip_fields}, can perform zero-shot object search tasks because CLIP is highly accurate for zero-shot classification tasks.
However, as explained in Section~\ref{sec:related:contrastive}, CLIP is poor at fine-grained classification.
Therefore, the semantic map created using CLIP feature vectors is unsuitable for InstanceImageNav.
Shafiullah~\textit{et~al.} suggested that the location at which a given image query is captured using CLIP-Fields can be coarsely identified.
However, whether the location of a specific object as required by InstanceImageNav can be searched has not been verified.



\begin{figure*}[!t]
    \centering
    \includegraphics[width=\linewidth]{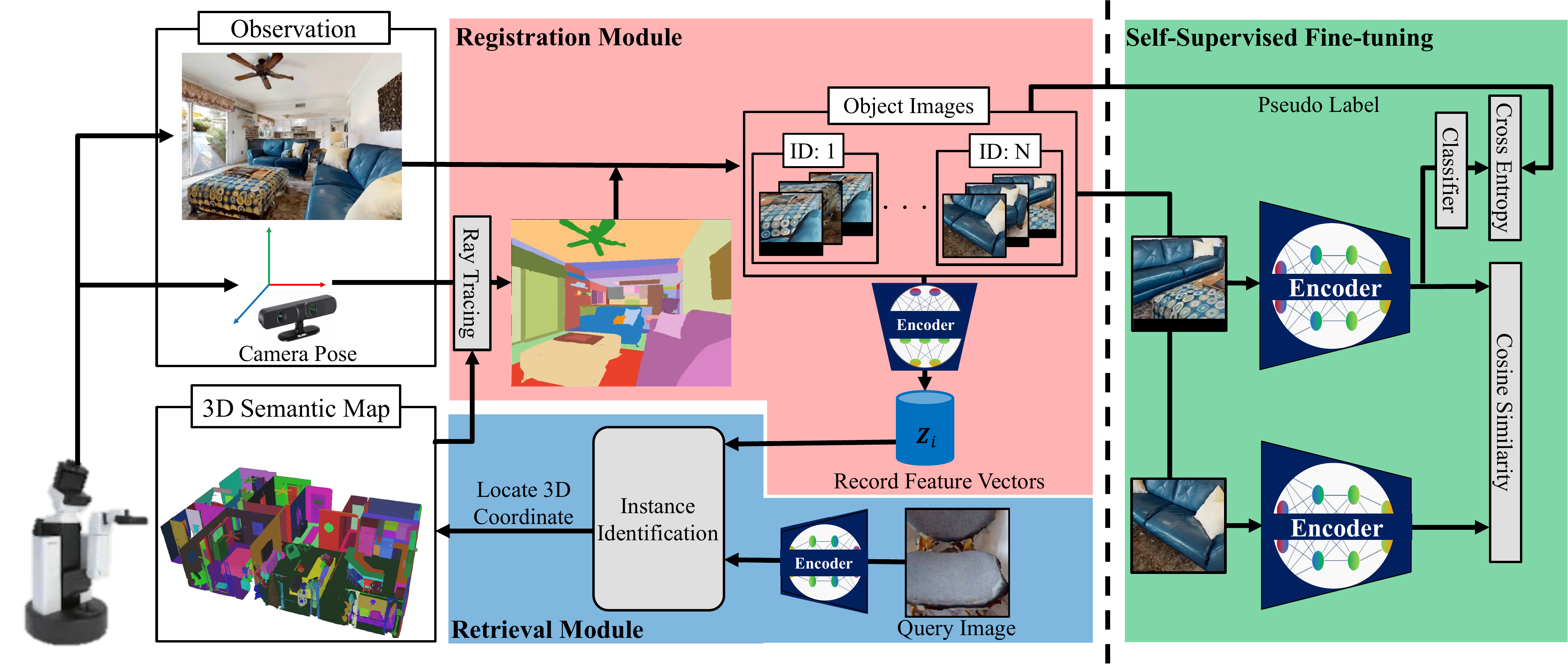}
    \caption{An overview of the proposed method. The red area indicates the recording of feature vectors for each object, whereas the blue area indicates performing a similar object search.}
    \label{fig:method}
\end{figure*}


\section{Proposed System}
\label{sec:proposed}


In the proposed system, a robot explores the environment, identifies an instance identical to a given query image among the collected object images, and uses a 3D semantic map of the environment to locate the position of the target object. 
In addition, we propose a method called \textbf{ semantic instance multi-view contrastive fine-tuning (SimView)} for fine-tuning pre-trained models using a self-supervised learning framework to improve the task accuracy in the environment.
Fig.~\ref{fig:method} shows an overview of the proposed system.

\subsection{Vector Registration using the Registration Module}
\label{sec:proposed:registration}

\subsubsection{Exploration of Environment}

Exploration points are set up at the same intervals in the free space on a 2D map, and the robot explores the environment by visiting these exploration points. 
In this study, the exploration points were placed at 30-cm intervals.

\subsubsection{Observation of Object Image}
\label{obs-module}

While the robot explores the environment as shown in Fig.~\ref{fig:method}, 2D mask images are generated using ray tracing with a segmented 3D map and camera poses.
Generating mask images from a 3D map enables the same object to be associated across different frames. 
The observed mask images are converted into bounding boxes (BBoxes), and the robot extracts the BBoxes regions from the observed RGB images to observe the objects. 
When the BBoxes of each object are extracted from the RGB image, it is adjusted to match the longer side of the BBox. 
In addition, areas outside the original RGB image are interpolated in only black.

\subsubsection{Registration of Feature Vector of Object Image}
The images of the observed objects are pre-processed and fed into a pre-trained encoder to convert them into feature vectors.
In this study, the observed images were resized to $256\times256$; subsequently, a $224\times224$-pixel image was cropped around the image center and normalized before being input into the encoder. 
The types of transformations used for pre-processing and the parameters used for normalization were the same as those employed in previous studies~\cite{simsiam}. 
In addition, when the same object was observed multiple times in the environment, the observed feature vectors for each object were recorded in a single set.

\subsection{Self-Supervised Fine-Tuning using an Instance Classifier}
\label{sec:proposed:fine-tuning}

This module fine-tunes the image encoder, which is pre-trained via contrastive learning using self-supervised learning, object images observed by the robot while exploring the environment, and their pseudo-labels.
When a robot explores an environment and observes objects, images of the same instance include those observed from various angles.
In a preliminary experiment, we confirmed that, when fine-tuning a pre-trained model using only contrastive learning on such a dataset, the accuracy of discrimination between instances is worse than that of the trained model.
Therefore, we propose a method for simultaneously training linear classifiers using contrastive learning. 
We use the object instance ID $y_{true}$ obtained from a 3D semantic map of the robot's environment as the pseudo-labels.

In addition, a contrastive learning method using negative pairs is recommended for training with a very large batch size and requires a large amount of data for learning~\cite{simclr}.
To conduct fine-tuning, the robot must continue exploring the environment for a long time and collect images of the objects.
Therefore, we use SimSiam for fine-tuning, which enables learning even with a small batch size~\cite{simsiam}.
This module conducts fine-tuning to minimize the following loss function:
\begin{equation}
    \begin{split}
        \mathcal{L} &= \frac{1}{2}\{\text{CosSim}(h, z^\prime) + \text{CosSim}(h^\prime, z)\} \\
        &\quad + \frac{1}{2}\{\text{CE}(y_{pred}, y_{true}) + \text{CE}(y^\prime_{pred}, y^\prime_{true})\},
    \end{split}
\end{equation}
where CE() and CosSim() denote cross-entropy and cosine similarity, respectively.
$z$ and $h$ are the outputs of SimSiam's projector and the predictor of the object image $x$ observed by the robot, respectively.
$y_{pred}$ is the instance ID of $x$ predicted by the classifier.
Here, $x$ and $x^{\prime}$ are different images of the same instance generated via data augmentation, respectively.
The same applies to the other primed variables, $z^{\prime}$, $h^{\prime}$, $y_{pred}^{\prime}$, and $y_{true}^{\prime}$.



\begin{figure}[t]
    \centering
    \includegraphics[width=\linewidth]{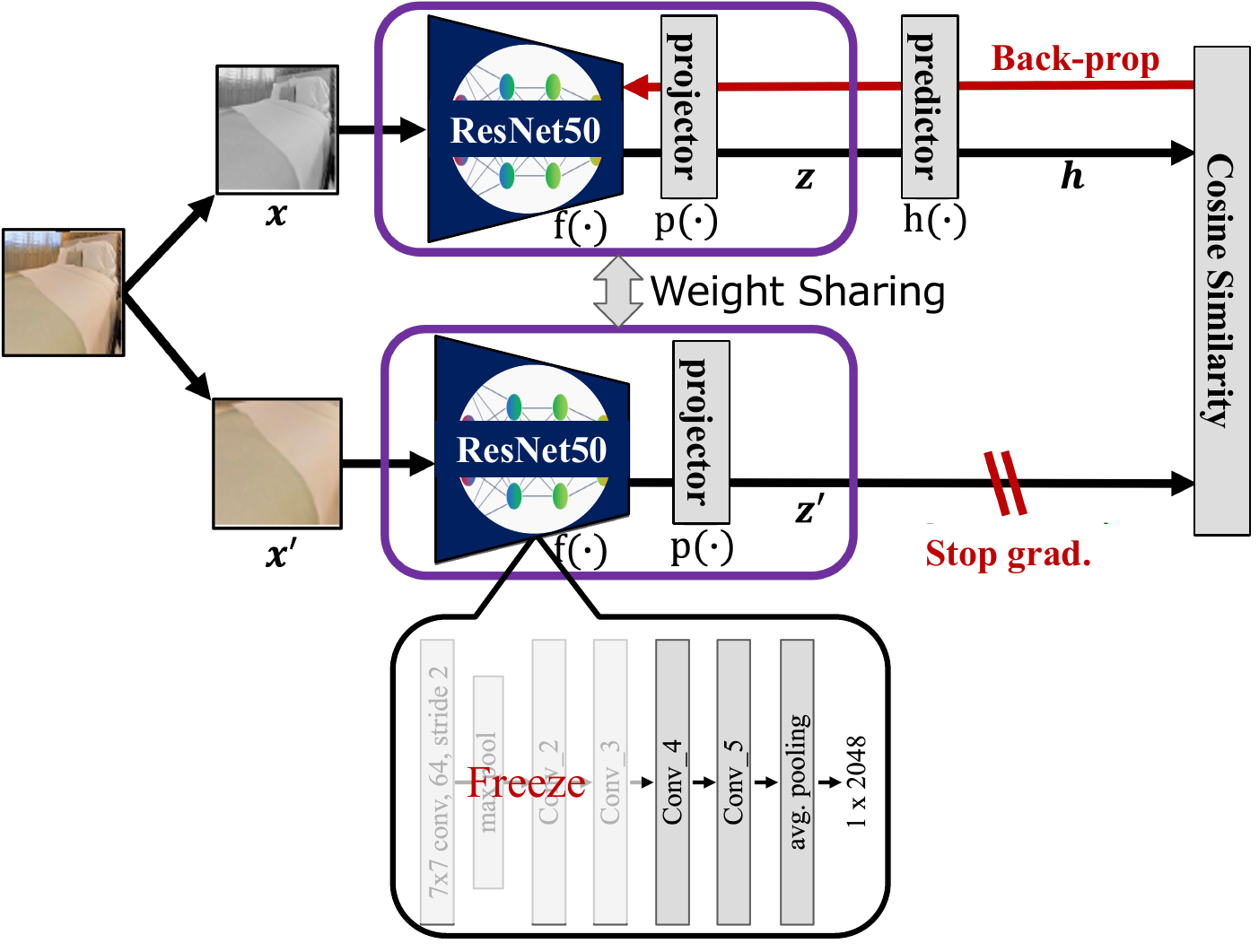}
    \caption{Parameters to be fixed during fine-tuning. All parameters up to the third block of ResNet50 used in SimSiam were fixed.}
    \label{fig:fine_tuning_detail}
\end{figure}

In addition, we can quickly introduce the robot into a task environment by reducing the time required for fine-tuning.
Therefore, we used a method to reduce the time required for fine-tuning by fixing certain parameters.
Fig.~\ref{fig:fine_tuning_detail} shows the relevant parts of the parameters that were fixed during fine-tuning.
Specifically, the parameters up to the third block of ResNet50~\cite{resnet} used in SimSiam were fixed and the other parameters were fine-tuned.

Please refer to the Appendix for details regarding the effectiveness of training SimSiam with a classifier.

\subsection{Instance Identification using the Retrieval Module}
\label{sec:proposed:retrieval}
The given query image $I_q$ is input into a pre-trained encoder, resulting in a feature vector denoted as $\bm{q}$.
Furthermore, we represent the observed feature vectors with instance ID $i$ as a set $\bm{Z}_i = \{\bm{z}_{i,n}\}_{n=1}^{N_i}$.
$N_{i}$ denotes the number of times an object with instance ID $i$ is observed.

The cosine similarity between the query image $\bm{q}$ and observed feature vectors $\bm{Z}_i$ is calculated as $\text{CosSim}(\bm{z}_{i, n}, \bm{q}) = \frac{\bm{z}_{i, n} \cdot \bm{q}}{\lVert\bm{z}_{i, n}\rVert \lVert\bm{q}\rVert}$ for each set of feature vectors corresponding to each instance ID, as follows:
\begin{align}
    \label{eq:similarity}
    m_i = {\rm{max}}(\{\text{CosSim}(\bm{z}_{i, n}, \bm{q})\}_{n=1}^{N_i}) .
\end{align}
From multiple similarities, the maximum value $m_i$ is selected.
An image encoder trained using contrastive learning embeds an image onto the surface of a spherical space~\cite{understanding_contrastive_learning}.

Finally, the instance ID $J_{\text{target}}$ with the highest similarity among the sets of maximum similarities $\{m_{j}\}_{j=1}^J$ for each instance is obtained as follows:
\begin{equation}
    \label{eq:argmax-sim}
    J_{\text{target}} = {{\argmax_{j}}}(\{m_j\}_{j=1}^J) .
\end{equation}
The position of the target object is obtained using the instance ID $J_{\text{target}}$ from the search results and the 3D semantic map.
The robot can navigate to a target position using a map.


\section{Experiment}

This experiment aimed to verify that an image encoder pre-trained by contrastive learning of only image pairs is more suitable than CLIP for identifying the same object as the query image, and to verify the effectiveness of fine-tuning using images observed by the robot.

\begin{table*}[t]
    \centering
    \caption{Results of Averaging mAP in Each Environment from 10 Repeated Trials of Similar-Image Search}
    \begin{tabularx}{1.0\linewidth}{|c|c|C|C|C|C|C|C|C|C|C|C|}
        \hline
        \textbf{Comparison Methods} & \textbf{Arch.} &  \textbf{Env. 1} & \textbf{Env. 2} & \textbf{Env. 3} & \textbf{Env. 4} & \textbf{Env. 5} & \textbf{Env. 6} & \textbf{Env. 7} & \textbf{Env. 8} & \textbf{Env. 9} & \textbf{Avg.} \\
        \hline
        \hline
        SimView (Ours) & ResNet50 & $\underline{\bm{0.79}}$ & $\underline{\bm{0.67}}$ & $\underline{\bm{0.68}}$ & $\underline{\bm{0.91}}$ & $\underline{0.68}$ & $\underline{\bm{0.45}}$ & $\underline{\bm{0.72}}$ & $\underline{\bm{0.74}}$ & $\underline{\bm{0.8}}$ & $\underline{\bm{0.72}}$ \\
        \hline
        SimSiam~\cite{simsiam} & ResNet50 & $\underline{\bm{0.70}}$ & $\underline{0.58}$ & $\underline{0.56}$ & 0.80 & 0.65 & $0.41$ & 0.63 & $\underline{0.68}$ & $\underline{0.75}$ & $\underline{0.64}$ \\
        DINOv2~\cite{dinov2} & ViT-B/14 & $0.51$ & $0.48$ & $0.45$ & $0.66$ & $\underline{\bm{0.71}}$ & $0.41$ & $0.55$ & $0.44$ & $0.61$ & 0.54 \\
        SimCLR~\cite{simclr} & ResNet50 & 0.65 & 0.56 & 0.52 & $\underline{0.83}$ & $0.64$ & $\underline{\bm{0.45}}$ & $\underline{0.66}$ & 0.62 & $\underline{0.75}$ & 0.63 \\
        \hline
        $\checkmark$CLIP~\cite{clip} & ResNet50 & 0.51 & 0.36 & 0.35 & 0.51 & 0.48 & 0.44 & 0.42 & 0.44 & 0.59 & 0.46 \\
        $\checkmark$CLIP~\cite{clip} & ViT-B/16 & 0.56 & 0.38 & 0.37 & 0.47 & 0.48 & 0.37 & 0.38 & 0.35 & 0.48 & 0.43 \\
        \hline
        \multicolumn{12}{p{0.9\linewidth}}{
            \vspace{0.5pt}
            The methods marked with $\checkmark$ represent those pre-trained through contrastive learning between text and images.
        }
    \end{tabularx} 
    \label{tab:result}
\end{table*}

\subsection{Dataset}

\begin{figure}[t]
    \centering
    \includegraphics[width=0.9\linewidth]{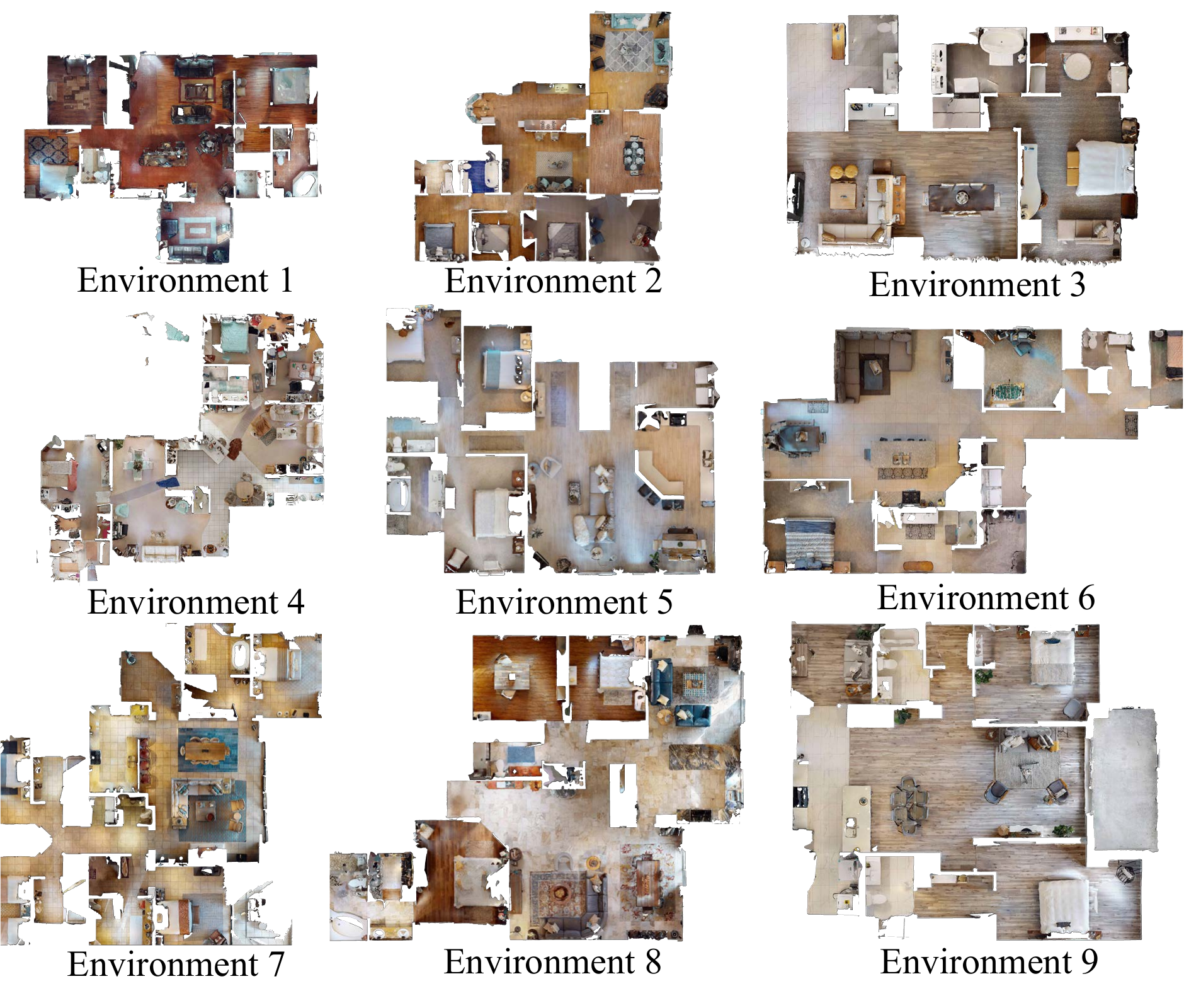}
    \caption{Allocentric view of the nine environments used in the experiment}
    \label{fig:exp_env}
\end{figure}


\begin{figure}[t]
    \centering
    \begin{tabular}{cc}
      \begin{minipage}[b]{0.47\linewidth}
        \centering
        \includegraphics[keepaspectratio, width=0.9\linewidth]{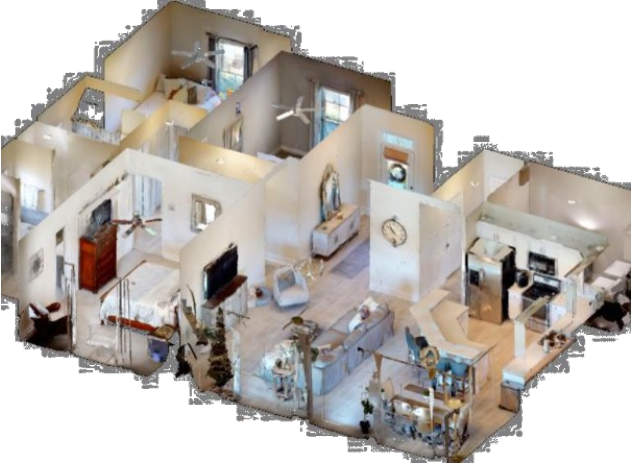}
        \subcaption{RGB 3D mesh}
      \end{minipage}
      \begin{minipage}[b]{0.47\linewidth}
        \centering
        \includegraphics[keepaspectratio, width=0.9\linewidth]{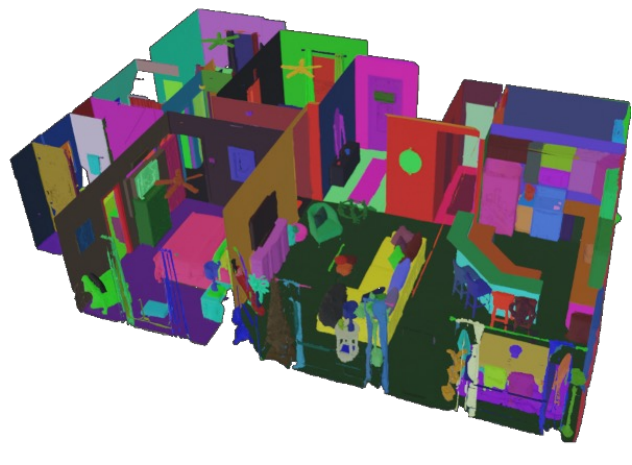}
        \subcaption{Semantic 3D mesh}
      \end{minipage}
    \end{tabular}
  \caption{Example 3D mesh data which is included in HM3D.}
  \label{fig:hm3d_example_3dmesh}
\end{figure}

We used nine 3D meshes from the Habitat Matterport 3D (HM3D) created by scanning actual home environments~\cite{hm3d}. 
These nine 3D meshes were scans of single-floor domestic environments, as shown in Fig.~\ref{fig:exp_env}. 
As shown in Fig.~\ref{fig:hm3d_example_3dmesh}, HM3D includes 3D meshes containing RGB data and 3D meshes segmented for each instance. 
In this experiment, we used 3D meshes containing RGB data with the habitat simulator~\cite{habitat} as the simulation environment, whereas we used segmented 3D meshes as the 3D semantic map of the environment of the robot.

Furthermore, we used images included in the dataset proposed by Krantz~\textit{et al}.~\cite{instancenav} as the query images.
This dataset includes parameters such as the camera position, orientation, and resolution when capturing a particular instance. 
Based on this information, images were captured. Using a method similar to that described in Section~\ref{obs-module}, only the regions of interest for the target instances were cropped to collect query images.

In addition, this dataset contains query images for over 10 different instances for each environment, all of which belong to one of the six classes: ``bed,'' ``chair,'' ``TV,'' ``plant,'' ``toilet,'' or ``couch.''

\subsection{Experimental Settings}

Stochastic gradient descent (SGD) was used as the optimizer for all conditions.
During the training process, we unified several hyperparameters across different training conditions. The batch size was set to 32, with a learning rate of 0.07. We used a weight decay of $1.5 \times 10^{-6}$ and a momentum of 0.9. The training was conducted for 100 epochs. These hyperparameters were selected to ensure consistency and comparability of the experiments.
The learning rate was allowed to decay as the training progressed using a cosine scheduler~\cite{cosine_scheduler}.
Four types of data expansion were used during training: crop and resize, color jitter, grayscale, and horizontal flip.
A PC with an Intel Core i9-9900K and an Nvidia RTX 2080 was used in this experiment.


\subsection{Metrics}
We used the mean average precision (mAP), which was used in a similar image-retrieval benchmark~\cite{object_retrieval}, for the evaluation metrics.
The mAP is computed by averaging the average precision (AP) for each instance. 
The AP is calculated as $\text{AP} = \frac{1}{N}\sum_{n=1}^{N}\text{P}_n$, where $N$ is the number of instances, and $\text{P}_n$ is the precision. 
Because the number of images to be retrieved is one, $\text{P}_n$ is 1 if the retrieved image is the same as the query image, and 0 otherwise.
Subsequently, the mAP is calculated as the average AP when $K$ different query images are prepared for each instance.
Therefore, the mAP is calculated as follows:
\begin{equation}
    \label{eq:mAP}
    \text{AP} = \frac{1}{K}\sum_{k=1}^{K}\text{AP}_k,
\end{equation}
where $K$ is the number of query images for each instance; in this experiment, $K = 10$.

We also investigated the reasons for the search failures.
There are two possible types of search failures.
\begin{enumerate}
    \item An object in the query image and its class are the same, but a different instance ID is retrieved.
    \item A different object from the class of query image is retrieved.
\end{enumerate}
We investigated the items of the two types of search failures in terms of the failure rate using the assigned instance ID and object class labels from both the query image and the 3D semantic map.
Each rate was calculated by performing the task 10 times, calculating the average number of failed searches in each environment, and averaging the results across all environments.

\subsection{Comparison Methods}

We compared the self-distillation methods SimSiam~\cite{simsiam} (with pre-trained ResNet50) and DINOv2~\cite{dinov2} (with pre-trained ViT), a contrastive learning method with image pairs SimCLR~\cite{simclr} (with pre-trained ResNet50), and CLIP~\cite{clip} (with pre-trained ResNet50 and ViT).
In addition, we compared the pre-trained models with a fine-tuned model using SimView.


\section{Results}

\begin{table}[t]
    \centering
    \caption{
        Results of the Error Analysis of Search Rates
    }
    \begin{tabularx}{1.0\linewidth}{|C|C|C|}
        \hline
        \textbf{Methods} & \textbf{Diff. instance $\downarrow$} & \textbf{Diff. class $\downarrow$} \\
        \hline
        \hline
        SimView (Ours) & $\underline{\bm{0.11}}$ & $\underline{\bm{0.17}}$ \\
        \hline 
        SimSiam & $\underline{0.17}$ & $\underline{0.19}$ \\ 
        DINOv2 & $0.25$ & $0.20$ \\ 
        SimCLR & $0.18$ & $0.19$ \\
        \hline
        $\checkmark$CLIP [ResNet] & $0.18$ & $0.36$ \\ 
        $\checkmark$CLIP [ViT] & $0.19$ & $0.38$ \\
        \hline
        \multicolumn{3}{p{0.9\linewidth}}{
            \vspace{0.5pt}
            (Diff. instance) Different instances within the same class;
        } \\
        \multicolumn{3}{p{0.9\linewidth}}{
            (Diff. class) Object images with different classes.
        }
    \end{tabularx}
    \label{tab:error-result}
\end{table}

\begin{figure*}[t]
    \centering
    \includegraphics[width=\linewidth]{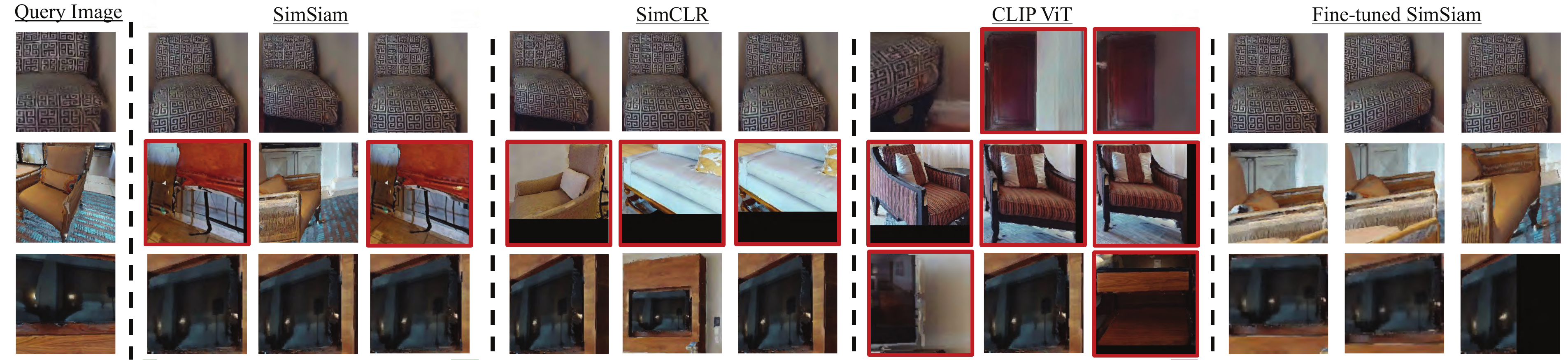}
    \caption{Results of obtaining the top three-nearest neighbors in the latent space of the query image. The red borders indicate images that contained a different instance from the query image.}
    \label{fig:neighbor-sampling}
\end{figure*}

\begin{figure*}[t]
    \centering
    \begin{tabular}{cc}
      \begin{minipage}[b]{0.48\linewidth}
        \centering
        \includegraphics[keepaspectratio, width=\linewidth]{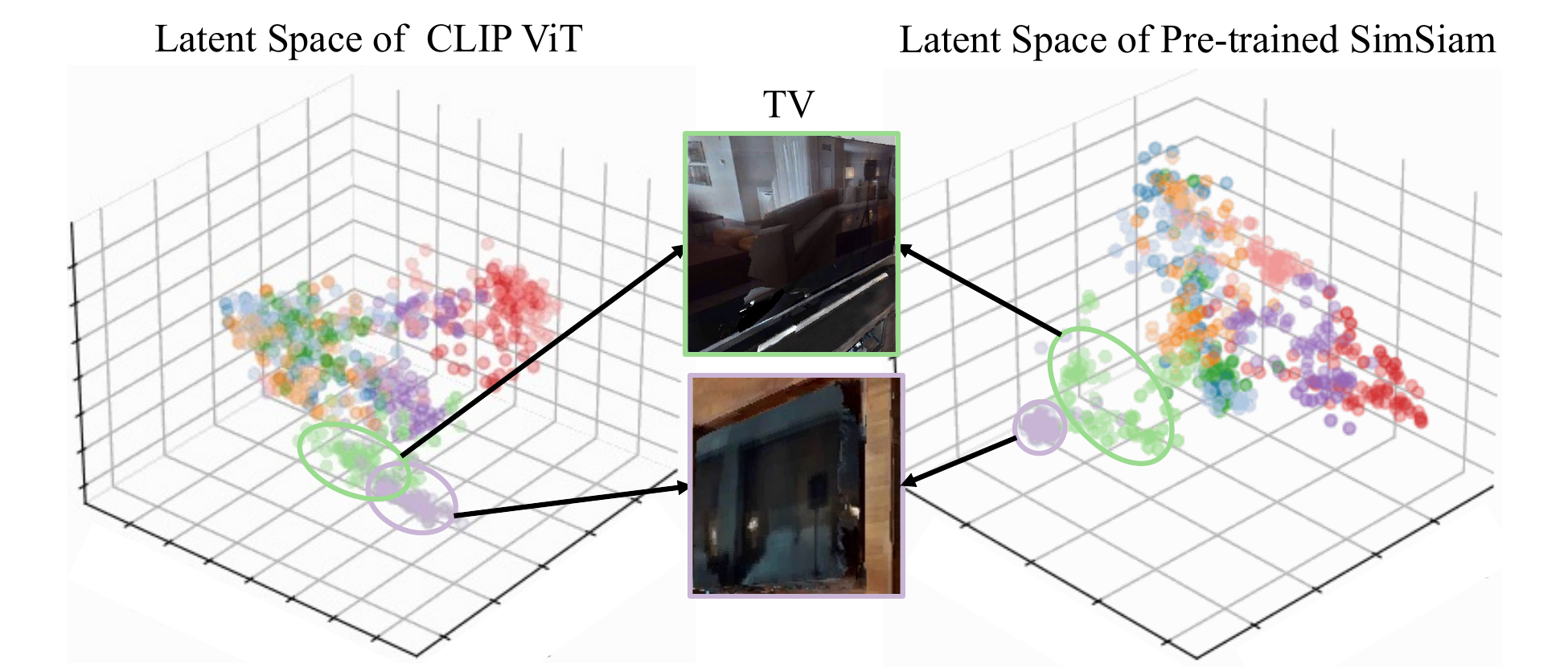}
        \subcaption{Comparison CLIP and ImageNet-pre-trained SimSiam}
      \end{minipage}
      \begin{minipage}[b]{0.48\linewidth}
        \centering
        \includegraphics[keepaspectratio, width=\linewidth]{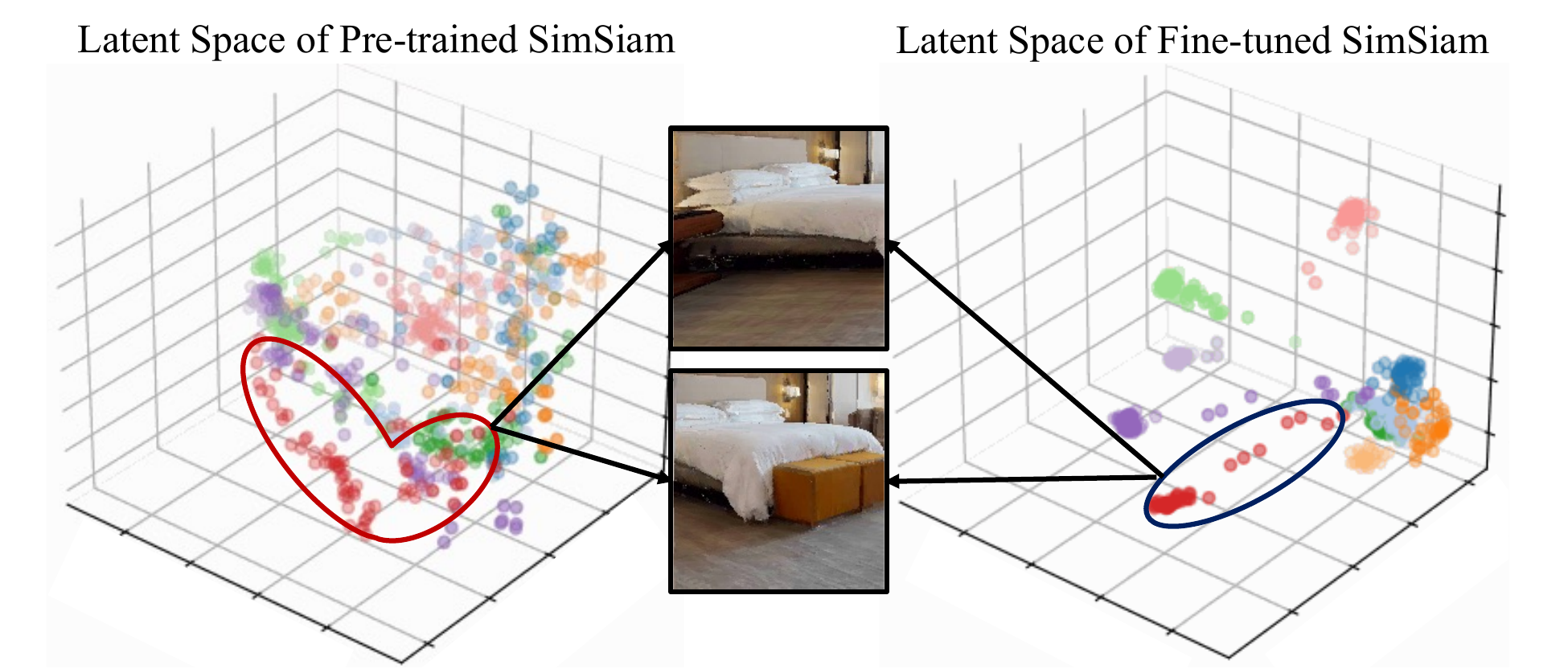}
        \subcaption{Comparison ImageNet-pre-trained and SimView}
      \end{minipage}
    \end{tabular}
  \caption{Latent spaces of image encoders on Env. 3. 
    In the scatter plot, different colors represent different instances. 
    In (a), the latent space of SimSiam pre-trained on ImageNet is more separated than the latent space of CLIP for images of different instances within the same class ``TV''. 
    In the red ellipse of (b), images of the same instance are scattered in the latent space of the pre-trained model. 
    However, In the blue ellipse of (b), the latent space of SimView more closely clustered images of the same instance than the pre-trained model.
    In (a) and (b), the latent space of the pre-trained SimSiam is the same, but the perspective of the 3D scatter plot differs between the two figures.
  }
  \label{fig: latent_space_comparison}
\end{figure*}

\subsection{Comparison Unimodal Contrastive Learning and CLIP}
Table~\ref{tab:result} lists the average mAP over 10 trials of similar image retrievals. 
SimSiam resulted in higher mAP values than the other methods in many environments.
In addition, SimCLR, a contrastive learning method other than SimSiam, resulted in higher mAP values in many environments. 
SimCLR was trained to maximize the similarity between two images generated through data augmentation and minimize the similarity with other images, which might result in a higher similarity between images of the same object. 
However, the results of CLIP tended to show a larger difference in mAP between SimSiam and CLIP than between SimSiam and SimCLR. 
This was likely because CLIP, unlike unimodal contrastive learning, does not learn the similarities between different images of the same object.
Table~\ref{tab:error-result} lists the classification failure rates. 
As shown in Tab.~\ref{tab:error-result}, compared with ImageNet-pre-trained SimSiam and SimView, CLIP retrieved images with a different instance or class from the query image at a higher rate.
Furthermore,  Fig.~\ref{fig:neighbor-sampling} shows the three neighborhood images of the query image for each method.
As shown in the bottom row of Fig.~\ref{fig:neighbor-sampling}, where a TV image is given as the query, the neighborhood images retrieved by CLIP included the image from a different instance than the query image and the image different from ``TV.''
In contrast, all three images retrieved by SimSiam and SimCLR were the same as the query images. 
These observations suggest that encoders pre-trained with CLIP are less adept at capturing fine visual similarities than unimodal contrastive learning.
As shown in Fig.~\ref{fig: latent_space_comparison} (a),  in the latent space of ImageNet-pre-trained SimSiam, images of different instances belonging to the same class ``TV'' were separated in the latent space compared with the latent space of CLIP.
Based on these results, we consider that an image encoder pre-trained with the image contrastive task is effective in searching for an object identical to a query image in the environment.


\subsection{Comparison Pre-trained SimSiam and SimView}
We also compared ImageNet-pre-trained SimSiam~\cite{ImageNet} and SimSiam fine-tuned with images collected by exploring the environment.
As shown in Tab.~\ref{tab:result}, eight of the nine environments improved their metric scores, with scores, on average, approximately $8\%$ better than those of the pre-trained SimSiam.
Furthermore, as shown in Fig.~\ref{fig: latent_space_comparison}, images of the same instance were scattered in the latent space of ImageNet-pre-trained SimSiam. However, in the latent space of the fine-tuned SimSiam on Env3, images of the same instance were densely distributed.
Hence, compared with ImageNet-pre-trained SimSiam, the fine-tuned SimSiam latent space could separate different instances from each other.
We believe that this change in the distribution of images in the latent space led to an improvement in mAP.
As shown in the bottom rows of Fig.~\ref{fig:neighbor-sampling}, the three neighborhood images retrieved from SimSiam or SimCLR contained instance images different from the query image.
However, the three neighborhood images retrieved by fine-tuned SimSiam were the same as those of the query image.


\section{Conclusion}
\label{sec:conclusion}

We propose SimView, which leverages multi-view images based on a 3D semantic map of the environment and self-supervised learning by SimSiam to train an instance identification model on-site. 
This study investigated whether an encoder trained through contrastive learning between images is effective for a robot searching for an instance in an environment that matches an object in a given query image.
The results showed that the encoder trained through self-distillation and SimView is effective for InstanceImageNav tasks compared with CLIP encoders pre-trained through contrastive learning between text and images, which are used in previous methods~\cite{nlmap, clip_fields}. 



A limitation of this validation is that both the observation and query images used were rendered images from 3D meshes containing RGB data, which may have introduced domain-shift effects with the training dataset of the encoder. 
For this validation, we assumed that the 3D map was pre-created. 
Therefore, in the future, we will incorporate the 3D semantic mapping method~\cite{voxblox_plusplus,kanechika_3Dmap} and validate it in the real world.


\section*{Appendix: Fine-Tuning Method Verification}
\label{sec:appendix}

We evaluated the effectiveness of fine-tuning a pre-trained SimSiam model by comparing two approaches: with or without an attached linear classifier. 
We also evaluated the performance difference between tuning all encoder layers and tuning only a subset.

\subsection{Multi-View Datasets}
\label{sec:appendix:dataset}


\textbf{Multi-view Images of Rotated Objects (MIRO)~\cite{miro}}:
MIRO includes images in which only objects appear on white backgrounds.
This dataset contains 12 object classes with 10 instances for each class.
Each instance was obtained from 160 camera poses.
Thus, the entire dataset contains 19,200 images.
In this validation, half of the images in the dataset were used for training, and the remainder were used for testing.
Image labels were assigned to each instance.


\textbf{Object Pose Incariance (ObjectPI)~\cite{objectpi}}:
This dataset differs from MIRO in that the images contain backgrounds other than the target object, which are similar to the objects observed by the robot in its environment.
In this dataset, image labels are assigned to each instance.
Each instance contains eight images captured in different directions.
ObjectPI's training dataset contains 400 different instances.
In addition, the test dataset contains 100 types of instances.
Therefore, in the verification using ObjectPI, we fine-tuned SimSiam using 3,200 images and tested it using 800 images.

\subsection{Metrics}
\label{sec:appendix:metrics}
To evaluate the feature representation of the fine-tuned image encoder, we clustered the image feature representation obtained from the encoder using k-means and assigned a cluster index to each image.
We then used the adjusted rand index (ARI) to quantify the degree of agreement between the true label assigned to each image in the dataset and the index assigned using k-means.
A higher ARI score can be interpreted as a feature representation that discriminates between images of the same and different instances in the latent space.


\subsection{Training Conditions}
\label{sec:appendix:condition}

We fine-tuned ImageNet-pre-trained SimSiam under four conditions: (a) updating all parameters with a classifier, (b) updating partial parameters with a classifier, (c) updating all parameters without a classifier, and (d) updating partial parameters without a classifier. 
Conditions (a) and (c) were run for 100 epochs, whereas conditions (b) and (d) were run for 50 epochs. 
The hyperparameters were maintained consistent across all conditions, except for the number of epochs. Under conditions (b) and (d), the parameters up to the third block of ResNet50 used in SimSiam were fixed, and the remaining parameters were updated (see Section~\ref{sec:proposed:fine-tuning}).


\subsection{Results}
\label{sec:appendix:result}

\begin{table}[!t]
    \begin{center}
        \caption{
            ARI Scores in Each Training Condition
        }
        \label{tab:fine_tuning_result}
        \begin{tabularx}{1.0\linewidth}{|cc|CC|CC|}
            \hline
            \multicolumn{2}{|c|}{\textbf{Training Condition}} & \multicolumn{2}{c|}{\textbf{MIRO}} & \multicolumn{2}{c|}{\textbf{ObjectPI}} \\
            \textbf{Update param.} & \textbf{Classifier} & \textbf{Train} & \textbf{Test} & \textbf{Train} & \textbf{Test} \\
            \hline
            \hline
            All & \checkmark & $\underline{0.49}$ & $\underline{0.53}$ & 0.14 & 0.20 \\
            Partial & \checkmark  & $\underline{\bm{0.85}}$ & $\underline{\bm{0.59}}$ &  $\underline{\bm{0.50}}$ & $\underline{\bm{0.66}}$ \\
            All & --- & 0.31 & 0.35 & 0.10 & 0.59 \\
            Partial & --- & 0.34 & 0.38 & $\underline{0.48}$ & $\underline{0.59}$ \\
            \hline
            \multicolumn{2}{|c|}{Pre-trained SimSiam} & 0.31 & 0.33 & 0.42 & 0.58 \\
            \hline
            \multicolumn{6}{p{0.9\linewidth}}{
                \vspace{0.5pt}
                The training and test data were created using MIRO and ObjectPI.
            }
        \end{tabularx}
    \end{center}
\end{table}

\begin{figure}[t]
    \centering
      \begin{minipage}[b]{0.49\linewidth}
        \centering
        \includegraphics[keepaspectratio, width=0.62\linewidth]{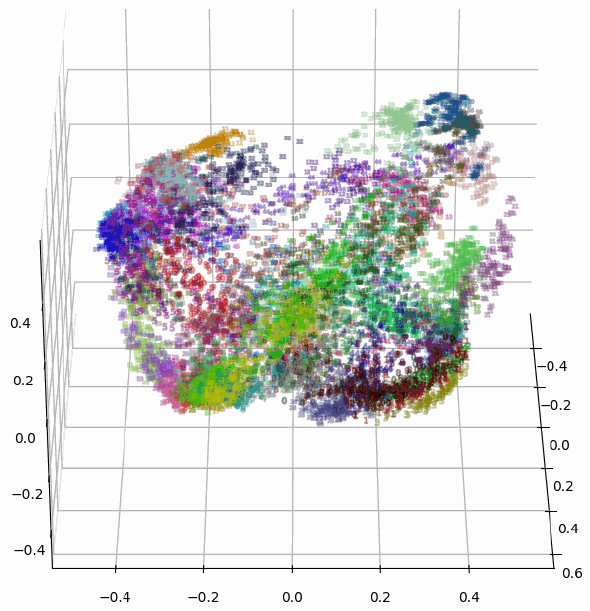}
        \subcaption{All params w/ classifier}
      \end{minipage}
      \begin{minipage}[b]{0.49\linewidth}
        \centering
        \includegraphics[keepaspectratio, width=0.62\linewidth]{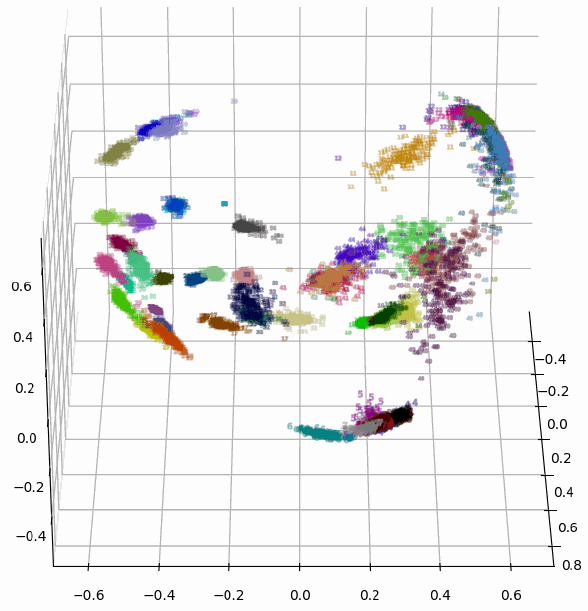}
        \subcaption{Partial params w/ classifier}
      \end{minipage} \\
      \begin{minipage}[b]{0.49\linewidth}
        \centering
        \includegraphics[keepaspectratio, width=0.62\linewidth]{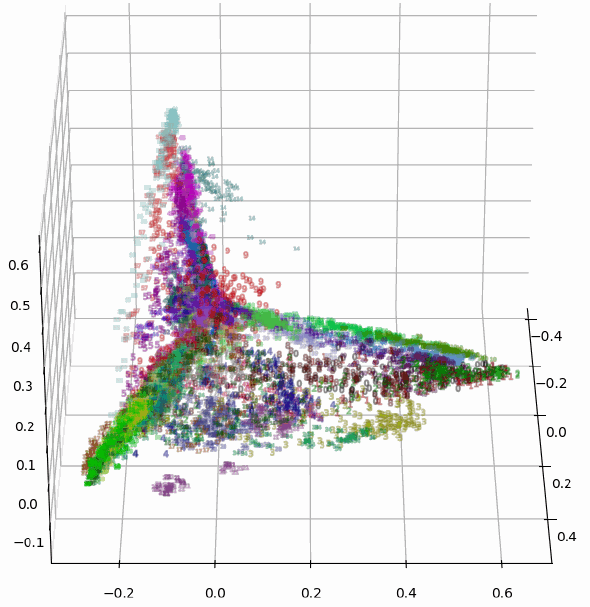}
        \subcaption{All params w/o classifier}
      \end{minipage}
      \begin{minipage}[b]{0.49\linewidth}
        \centering
        \includegraphics[keepaspectratio, width=0.62\linewidth]{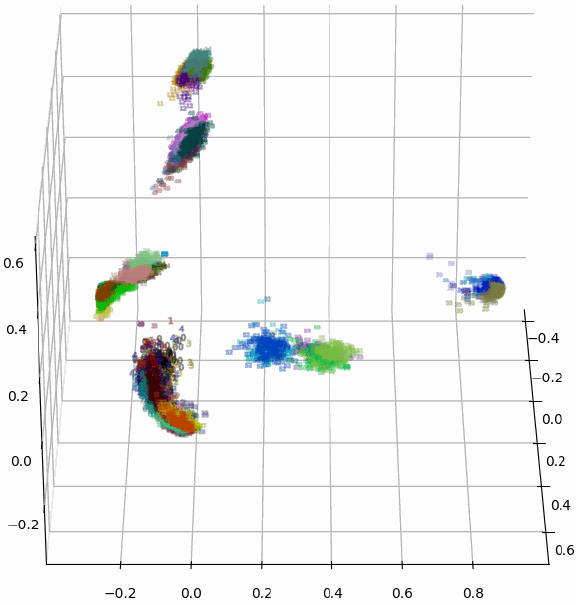}
        \subcaption{Partial params w/o classifier}
      \end{minipage}
      %
  \caption{Scatter plots visualizing the feature vectors calculated using the image encoder trained with the MIRO dataset for each condition, reduced in dimension using PCA. 
  Different colors represent different instances.}
  \label{fig:miro_latent_space}
\end{figure}


Table~\ref{tab:fine_tuning_result} lists the ARI scores under each learning condition for the training and test data generated using MIRO and ObjectPI.
The table shows that the condition where some parameters were learned using a linear classifier had the highest ARI score for any dataset.


Fig.~\ref{fig:miro_latent_space} shows the results of reducing the feature vectors of the images of the MIRO training dataset, which were calculated using the fine-tuned image encoder under each training condition, to three dimensions using principal component analysis (PCA). 
When training with a linear classifier, images are distributed throughout the latent space.
Furthermore, when some parameters are updated with the attached classifier, images of different instances can be separated in the latent space of each instance compared with when all the parameters are updated.
Therefore, the ARI score is expected to be the highest when a linear classifier is attached and only partial parameters are updated.


However, when fine-tuning is conducted using only contrastive learning, images are distributed only in part of the latent space, as shown in Fig.~\ref{fig:miro_latent_space}.
This suggests that if fine-tuning is performed using only contrastive learning on a multiview dataset, learning discriminative feature representations is difficult.
This may be caused by the appearance of the same object in the multi-view dataset, which differs depending on the angle of view when the object is captured.

Based on these results, an effective method to fine-tune the pretrained SimSiam is by adding a linear classifier to train SimSiam in the fine-tuning module of the proposed system.




\bibliographystyle{templates/IEEEtran}
\bibliography{root}




\end{document}